\documentclass[letterpaper]{article} 
\usepackage{aaai23}  
\usepackage{times}  
\usepackage{helvet}  
\usepackage{courier}  
\usepackage[hyphens]{url}  
\usepackage{amsfonts}
\usepackage{booktabs}
\usepackage{graphicx} 
\usepackage{natbib}  
\usepackage{caption} 
\usepackage{algorithm}
\usepackage{algorithmic}

\usepackage{newfloat}
\usepackage{listings}
\DeclareCaptionStyle{ruled}{labelfont=normalfont,labelsep=colon,strut=off} 
\floatstyle{ruled}
\newfloat{listing}{tb}{lst}{}
\floatname{listing}{Listing}
\usepackage{amsmath}

\title{Towards Safe Mechanical Ventilation Treatment \\Using Deep Offline Reinforcement Learning}
\author{
Flemming Kondrup\equalcontrib\textsuperscript{\rm 1},
Thomas Jiralerspong\equalcontrib\textsuperscript{\rm 1},
Elaine Lau\equalcontrib\textsuperscript{\rm 1},
Nathan de Lara\textsuperscript{\rm 1},\\
Jacob Shkrob\textsuperscript{\rm 1},
My Duc Tran\textsuperscript{\rm 1},
Doina Precup\textsuperscript{\rm 1,2},
Sumana Basu\textsuperscript{\rm 1,2}
}

\affiliations {
\textsuperscript{\rm 1}McGill University\\
\textsuperscript{\rm 2}Mila\\
Correspondance to: \emph{flemming.kondrup@mail.mcgill.ca}
}

\begin{document}

\maketitle

\begin{abstract}
Mechanical ventilation is a key form of life support for patients with pulmonary impairment. Healthcare workers are required to continuously adjust ventilator settings for each patient, a challenging and time consuming task. Hence, it would be beneficial to develop an automated decision support tool to optimize ventilation treatment. We present DeepVent, a Conservative Q-Learning (CQL) based offline Deep Reinforcement Learning (DRL) agent that learns to predict the optimal ventilator parameters for a patient to promote 90 day survival. We design a clinically relevant intermediate reward that encourages continuous improvement of the patient vitals as well as addresses the challenge of sparse reward in RL. We find that DeepVent recommends ventilation parameters within safe ranges, as outlined in recent clinical trials. The CQL algorithm offers additional safety by mitigating the overestimation of the value estimates of out-of-distribution states/actions. We evaluate our agent using Fitted Q Evaluation (FQE) and demonstrate that it outperforms physicians from the MIMIC-III dataset.
\end{abstract}

\section{Introduction}

The COVID-19 pandemic has put enormous pressure on the healthcare system, particularly on intensive care units (ICUs). In cases of severe pulmonary impairment, mechanical ventilation assists breathing in patients and acts as the key form of life support. However, the optimal ventilator settings are individual specific and often unknown \cite{Zein2016}, leading to ventilator induced lung injury (VILI), diaphragm dysfunction, pneumonia and oxygen toxicity \cite{Pham2017}. To prevent these complications, and offer optimal care, it is necessary to personalize mechanical ventilation.

Various efforts have proposed the use of machine learning (ML) to personalize ventilation treatments. These include the use of deep supervised learning \cite{Akbulut2014,Venkata2021} which permits high-level feature extraction, yet ignores the sequential nature of ventilation. Furthermore, supervised learning methods can only hope to imitate the physician's policy, which may lead to suboptimal treatment. Meanwhile, reinforcement learning (RL) interacts with the environment and gets immediate feedback from the patient in the form of rewards and hence can improve upon the physician's policy. Tabular RL has recently shown strong potential in mechanical ventilation \cite{Peine2021}, but, to the best of our knowledge, no previous works have attempted to combine deep learning and RL to improve mechanical ventilation.

We propose DeepVent, a Deep RL model to optimize mechanical ventilation settings and hypothesize it will lead to improved care. We consider both performance and patient safety with the aim of bridging the gap between research and real-life implementation. Here are our main contributions:

\begin{itemize}
\setlength\itemsep{0pt}
    \item We introduce DeepVent, a Deep RL model based on the Conservative Q-Learning algorithm \cite{kumar2020conservative}, and show using Fitted Q Evaluation (FQE) that it achieves higher performance when compared to physicians as recorded in the MIMIC-III dataset \cite{mimic}, behavior cloning and Double Deep Q-Learning (DDQN) \cite{vanhasselt2015deep}, a common RL algorithm in health applications.
    \item We compare DeepVent's decisions to those of physicians and of the DDQN agent. We show that DeepVent makes recommendations within safe ranges, as supported by recent clinical studies and trials. In contrast, DDQN makes recommendations in ranges unsupported by clinical guidelines. We hypothesize that this may be due to DDQN's overestimation of out-of-distribution states/actions and demonstrate the potential of Conservative Q-Learning to address this. This is essential in healthcare, where risk in decision making must be avoided.
    \item We introduce a clinically relevant intermediate reward applicable to many fields of healthcare. RL models can benefit highly from an intermediate reward, as it can permit faster convergence and improved performance \cite{Mataric}, and thus better outcomes for patients. Most previous efforts implementing RL in healthcare either did not address this or proposed a reward requiring important domain knowledge (see Section \ref{intermediateR}). Our intermediate reward is based on the Apache II mortality prediction score \cite{Apache2}, commonly used by physicians in ICUs, and leads to improved performance.
\end{itemize}

\section{Background \& Related Work}

\subsection{Reinforcement Learning (RL)}
RL is usually formalized as a \textit{Markov Decision Process} (MDP), which is defined by a tuple $(\mathcal{S}, \mathcal{A}, P, r, \gamma)$, where $\mathcal{S}$ is the state space, $\mathcal{A}$ the action space, $P$ the transition function defining the probability of arriving at a given state $s_{t+1}$ after taking action $a_t$ from state $s_t$, $r$ the reward function defining the expected reward received after taking action $a_t$ from state $s_t$ and $\gamma\in (0,1)$ the discount factor of the reward. At each time step $t$ of an episode, the agent observes the current state $s_t\in \mathcal{S}$, takes an action $a_t\in \mathcal{A}$, and transitions to another state $s_{t+1}\in \mathcal{S}$ while receiving a reward $r_t$. The goal of RL is to train a policy $\pi: \mathcal{S} \times \mathcal{A} \rightarrow [0,1]$ that maximizes the cumulative discounted return, $\sum_{t=0}^{T}\gamma^tr_t$ received over the course of an episode with $T$ timesteps.

\subsection{Q-Learning and Deep Q-learning}
Q-Learning \citep{qlearning} is one of the main RL algorithms and the most common method in healthcare applications \cite{RLHealthcareSurvey}. It aims to estimate the value of taking an action $a$ from a state $s$, known as the Q-value $Q(s,a)$. At each timestep $t$, upon taking action $a_t$ from state $s_t$ and transitioning to state $s_{t+1}$ with reward $r_t$, the agent updates the Q-value for $(s_t, a_t)$ as follows:\\
\begin{equation}\label{qvalueupate}
    Q(s_t, a_t)=Q(s_t, a_t) + \eta(r_t +\gamma \max _{a} Q(s_{t+1}, a) - Q(s_t, a_t))
  \end{equation}
where  $\eta\in(0,1)$ is the learning rate and $(r_t +\gamma \max _{a} Q(s_{t+1}, a))$ is the {\em target} of the update.
When the number of states is intractable, it becomes impractical to store in a table the $Q$-values for all state-action pairs. We can however use a function approximator to estimate the $Q$-values. The Deep Q Network (DQN) \citep{deepqnetworks} algorithm combines Q-Learning with deep neural networks to handle complex RL problems.
Despite offering many advantages, such as the ability to learn from data gathered through any way of behaving, and to generalize potentially to many states from a limited sample, DQN comes with challenges, such as the potential to substantially overestimate certain $Q$-values. Overestimation occurs when the estimated mean of a random variable is higher than its true mean. Because DQN updates its $Q$-values towards the target $r_t + \gamma\max_{a}Q(s_{t+1}, a)$, which includes the highest $Q$-value of the next state $s_{t+1}$, and because this is usually a noisy estimate, it can lead to an overestimation.

\subsection{Double Deep Q-Network (DDQN)} \label{sec:2.4}

DDQN \citep{vanhasselt2015deep} was introduced as a solution to the  overestimation problem in Q-learning. While DQN uses a single network to represent the value function, DDQN uses two different networks, parametrized by different parameter vectors, $\theta$ and $\theta^{\prime}$. At any point in time, one of the networks, chosen at random, is updated, and its target is computed using the $Q$-value estimated by the other network.
Thus, for network $Q_{\theta}$, the target of the update is:
\begin{equation}\label{new_target}
\begin{aligned}
r_t + \gamma Q_{\theta^{\prime}} (s_{t+1}, \arg\max_{a} Q_{\theta}(s_{t+1}, a))
\end{aligned}
\end{equation}
While this is beneficial, DDQN may still suffer from overestimation \cite{vanhasselt2015deep}, especially in offline RL.

\subsection{Offline Reinforcement Learning}
Traditional RL methods are based on an online learning paradigm, in which an agent actively interacts with an environment. This is an important barrier to RL implementation in many fields, including healthcare \citep{Levine2020}, where acting in an environment is inefficient and unethical, as it would mean putting patients at risk. Consequently, recent years have witnessed significant growth in offline (or batch) RL, where the learning utilizes a fixed dataset of transitions $\mathcal{D}=\left\{\left(s_t^{i}, a_t^{i}, r_t^{i}, s_{t+1}^{i}\right)\right\}_{i=1}^{N}$. 
Since the understanding of the environment of the RL model is limited to the dataset, this can lead to the overestimation of $Q$-values of state-action pairs which are under-represented in the dataset, or out-of-distribution (OOD). In the healthcare setting, this may translate to unsafe recommendations, putting patients at risk.

\subsection{Conservative Q-Learning (CQL)} \label{sec:2.6} 
Conservative Q-Learning (CQL) was proposed to address overestimation in offline RL \cite{kumar2020conservative}. It learns a conservative estimate of the $Q$-function by adding a regularizer $\label{conservativeminterm} \mathbb{E}_{\mathbf{s_t} \sim \mathcal{D}, \mathbf{a_t} \sim A}[Q(\mathbf{s_t}, \mathbf{a_t})]$ on the $Q$-learning error, in order to minimize the overestimated values of unseen actions. In addition, the term $-\mathbb{E}_{\mathbf{s_t},\mathbf{a_t} \sim \mathcal{D}}[Q(\mathbf{s_t}, \mathbf{a_t})]$ is added to maximize the $Q$-values in the dataset. In summary, CQL minimizes the estimated $Q$-values for all actions while simultaneously maximizing the estimated $Q$-values for the actions in the dataset, thus preventing overestimation of OOD or underrepresented state-action pairs.

\subsection{Related work}

\subsubsection{Algorithms for ventilation optimization}

Current approaches for ventilation optimization in hospitals commonly rely on proportional-integral-derivative (PID) control \citep{pid_article}, which are known to be sub-optimal \cite{pid_improved}. The use of more sophisticated machine learning methods have been suggested in recent years \citep{Akbulut2014,Venkata2021, pid_improved}. Recently, RL was proposed using a simple tabular approach \cite{Peine2021}. This was already expected to outperform clinical standards, providing strong evidence for the use of RL in this setting. Nonetheless, to the best of our knowledge, no Deep RL approach has been proposed for ventilation settings optimization. Furthermore, many core RL challenges, such as sparse reward and value overestimation, have not yet been addressed.

\subsubsection{Intermediate rewards in healthcare}
\label{intermediateR}

RL has been suggested in various fields of healthcare, such as sepsis treatment \cite{raghu2017deep,peng2019improving}, heparin dosage \citep{Heparin1}, mechanical weaning \citep{prasad2017reinforcement,Yu2020} and sedation \citep{Eghbali}. In RL, the use of a dense reward signal can help credit assignment \citep{Mataric}, leading to faster convergence and improved performance, which in healthcare translates to better outcomes for patients. Nonetheless, most previous attempts listed above either did not address this or proposed a reward requiring important domain specific knowledge. There is therefore an important need to develop intermediate rewards that both perform well and are broadly applicable to various fields of healthcare.

\section{Methods}

\begin{figure}[ht!]
\begin{center}
    \includegraphics[width=\columnwidth]{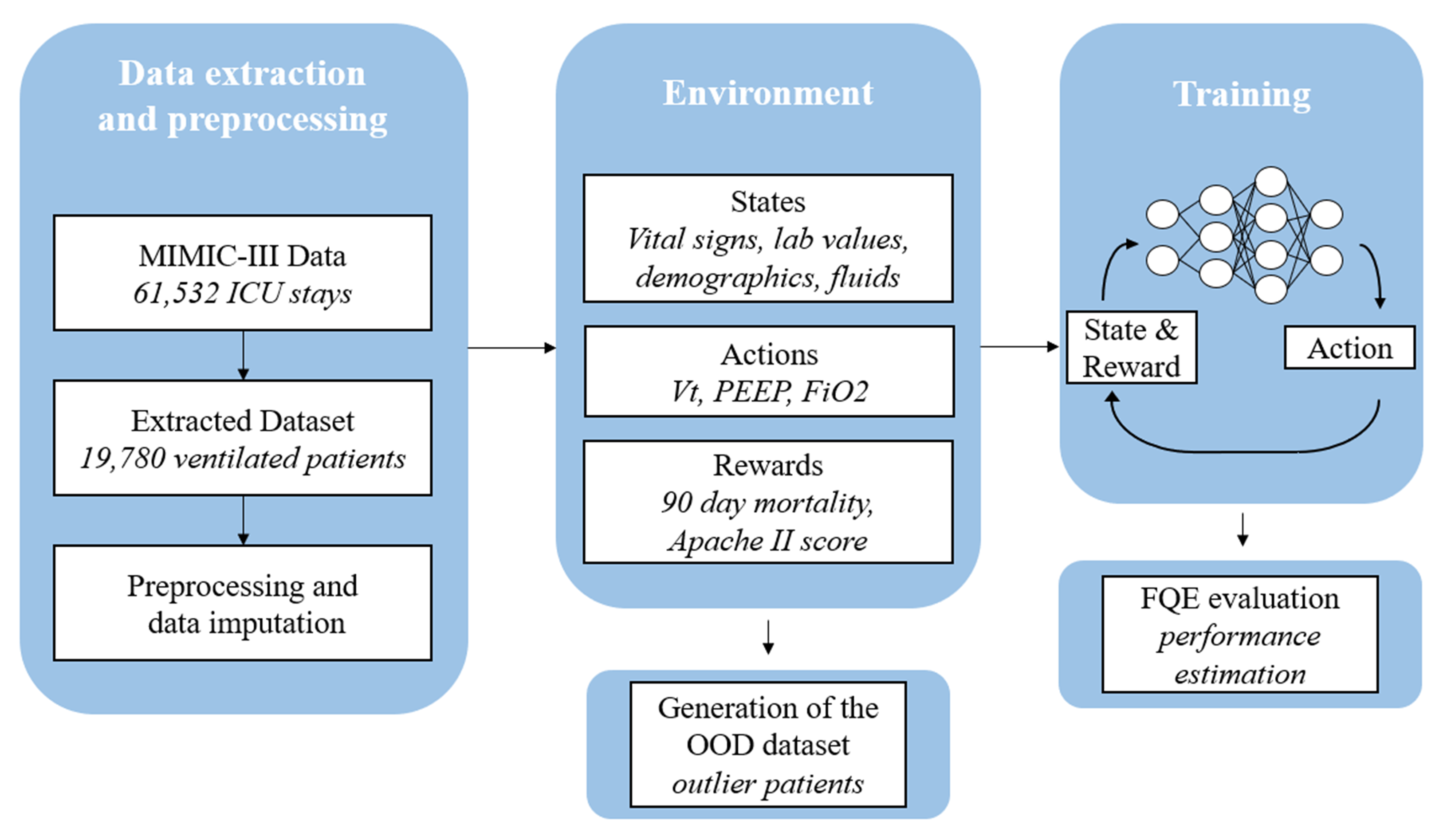}
    \caption{Overview of methods pipeline}
    \label{fig:pipeline}
\end{center}
\end{figure}

This section covers our methods, from data extraction and preprocessing to defining the RL problem, the generation of an out-of-distribution (OOD) dataset and our experimental setup. An overview of the pipeline is here in Figure \ref{fig:pipeline}.

\subsection{Data extraction and pre-processing}
\label{pre}

We used the MIMIC-III database \citep{mimic}, an open-access database containing data for 61,532 ICU stays at the Beth Israel Deaconess Medical Center (Boston, MA, USA) between 2001 and 2012. Standardized Query Language (SQL) was used to extract patient data into a table of four-hour time windows. For each patient, the following data were extracted: vital signs, lab values, demographics, fluids and ventilation settings. The first 72 hours of ventilation were selected. The patient data was separated into parallel state, action and reward arrays. For data imputation, a mix of methods was used. If less than 30\% of the data was missing, $k$-nearest-neighbor (KNN) imputation was used with $k=3$ \citep{missingdata}. For 30\% to 95\%, a time-windowed sample-and-hold method was used, whereby we took the initial value and used it to replace the following values, until either a new value was met or a limit was reached \citep{missingdata}. When the initial value was missing, mean value imputation was performed. Finally, for over 95\%, the variable was removed from our state space.

\subsection{RL Problem Definition}

Our MDP is defined similarly to the work of \cite{Peine2021}, with episodes lasting from the time of the patient's intubation to 72 hours afterwards.

\textbf{State Space} The state space $\mathcal{S}$ comprises 36 variables\footnote{For variable definitions refer to \citep{mimic}}:
\begin{itemize}
\setlength\itemsep{0pt}
\item Demographics: Age, gender, weight, readmission to the ICU, Elixhauser score
\item Vital Signs: SOFA, SIRS, GCS, heart rate, sysBP, diaBP, meanBP, shock index, temperature, spO2
\item Lab Values: Potassium, sodium, chloride, glucose, bun, creatinine, magnesium, carbon dioxide, Hb, WBC count, platelet count, ptt, pt, inr, pH, partial pressure of carbon dioxide, base excess, bicarbonate
\item Fluids: Urine output, vasopressors, intravenous fluids, cumulative fluid balance
\end{itemize}

\textbf{Action Space} The 3 ventilator settings of interest are:
\begin{itemize}
\setlength\itemsep{0pt}
\item Ideal weight adjusted tidal volume \textit{or} Vt (Volume of air in and out with each breath adjusted by ideal weight)
\item PEEP (Positive End Expiratory Pressure)
\item FiO2 (Fraction of inspired oxygen)
\end{itemize}

The action space $\mathcal{A}$ is the Cartesian product of the set of these three settings. Each setting can take one of seven values corresponding to ranges. We thus have an action as the tuple \(a = (v, o, p)\) with \(v \in Vt, o \in FiO_2,  p \in PEEP \).

\textbf{Reward Function} The main objective of our agent is to keep a patient alive long-term. Therefore, even if DeepVent only treats patients for 72 hours, it learns how to maximize their 90 day survival. This permits us to not only consider patient welfare during treatment but additionally prevent complications with long-term effects. We thus define a terminal reward $r(s_t, a_t,s_{t+1})$, which takes at the final state of an episode the value $-1$ if the patient passes away within $90$ days and $+1$ otherwise.
Because the sole use of a sparse terminal reward is known to cause poor performance in RL tasks \citep{Mataric}, we developed an intermediate reward based on the Apache II score \citep{Apache2}, which is widely used in ICUs to assess the severity of a patient's disease. The APACHE-II score compiles various physiological variables and determines how far from the healthy range a patient is. In order to reduce any source of bias, we made modifications to the APACHE II score to shape our reward function. We removed the FiO2 and respiratory rate (RR) variables as including them may have favored giving a ”normal range” FiO2 or RR. However, changing these variables may be required in certain cases. For example, it is well known that hypoxic patients can heavily benefit from a momentary increase in FiO2. In addition, the hematocrit variable was removed due to a high level of missingness (see Section \ref{pre}). Our modified score therefore contains the following variables: temperature, mean BP, heartrate, arterial pH, sodium, potassium, creatinine, WBC and GCS score (see \cite{Apache2} for more details). Since each variable in the APACHE II score contributes independently to the final score \citep{Apache2}, removing some of the variables does not detract from the ability of other variables to provide us with an indication of the gravity of a patient’s condition. In order to not simply define the reward based on how well a patient is doing but rather their evolution through time, our intermediate reward consists of the change in Apache II score between $s_{t+1}$ and $s_t$, which is normalized by dividing it by the total range of the score. Combining the intermediate and terminal rewards, we obtain our final reward function:

\[  
r(s_t^i, a_t^i, s_{t+1}^i) = 
     \begin{cases}
       +1 \quad\text{if $t+1=l_i$ and $m_{t+1}^i=1$}\\
       -1 \quad\text{if $t+1=l_i$ and $m_{t+1}^i=0$}\\
       \frac{(A_{t+1}^i - A_t^i)}{\text{max}_A - \text{min}_A} \quad\text{otherwise}\\
     \end{cases}
\]
where:
\begin{description}
\setlength\itemsep{0pt}
\item[$A_t^i$] is the modified Apache II score of patient $i$ at timestep $t$
\item[$m_t^i$] = 0 if patient $i$ is dead at timestep t and $1$ otherwise
\item[$l_i$] is the length of patient $i$'s stay at the ICU
\item[$\text{max}_A, \text{min}_A$] are respectively the maximum and minimum possible values of our modified Apache II score
\end{description}

\subsection{Generation of the Out-of-distribution dataset}
\label{ood_dataset}
To investigate the overestimation of DeepVent and DDQN, an out-of-distribution (OOD) set of patients was created. An outlier patient was defined as having at least one state feature (demographic, vital sign, lab value or fluid) at the beginning of ventilation in the top or bottom 1\% of the distribution. Approximately 25\% of patients were considered outliers.

\subsection{Experimental setup}

\subsubsection{Baselines}

We utilize three baseline methods: the \textit{physician policy}, \textit{Behavior Cloning (BC)} and a \textit{DDQN} model. The physician policy is the combination of all the transitions ($s_t, a_t, s_{t+1}$) found in the dataset. As such, it represents the choices made by the physicians attending to MIMIC-III patients. BC aims to predict physician choices using a supervised learning approach. It does so by training a policy network $\pi_{\theta}$ with parameter $\theta$ to predict the action $a_t$ taken by the physician based on the current state $s_t$ through the minimization of the categorical cross-entropy loss function $L(\theta) = E_{a_t,s_t \sim D}[-\sum_a p(a|s_t)\text{log}\pi_\theta(a|s_t)]$, where $p(a|s_t) = 0, \forall a\neq a_t$ where  $a_t$ is the action taken by the physician at  $s_t$. BC thus serves as a benchmark for non-RL methods. Finally, we use a DDQN model to serve as a Deep RL baseline. Our implementation of CQL is built on top of our DDQN, permitting  easy evaluation of the utility of adding the conservative aspect. DDQN and CQL are implemented using the d3rlpy library \citep{seno2021d3rlpy}.

\subsubsection{Training and Hyperparameters} 

Patient episodes were split into training (80\%) and validation (20\%). A grid search was conducted for the learning rate $\eta$, the discount factor $\gamma$ and the scaling factor $\alpha$ of the conservative effect of CQL. We considered $\eta$ values in $[1^{-7}, 1^{-6}, 1^{-5}, 1^{-4}]$, $\gamma$ values in $[0.25, 0.5, 0.75, 0.9, 0.99]$ and $\alpha$ values in [0.05, 0.1, 0.5, 1, 2]. Furthermore, the sigmoid and ReLU functions and architectures of 1 to 3 hidden layers of 64, 128, 256 and 512 nodes each were investigated. We trained these architectures for 1 million steps and determined optimal values of $\gamma=0.75$ and $\eta=1^{-6}$  for DDQN, and $\gamma=0.75$, $\eta=1^{-6}$ and $\alpha=0.1$ for CQL. The best architecture had 2 hidden layers with 256 units each and the ReLU function. 5 runs of 2 million steps were then performed and averaged for our results.

\subsubsection{Off-Policy Evaluation}
\label{OPE}
In online RL, policies are typically evaluated through interaction with the environment. However, in the healthcare setting where the environment is real patients, evaluating the policies in this manner would be unsafe. Evaluation is therefore done by using the dataset through methods grouped under the term Off-Policy Evaluation (OPE). 
The performance of these methods was recently evaluated in the healthcare setting \citep{tang2021model}, where Fitted Q Evaluation (FQE) \citep{FQE} consistently provided the most accurate results. Following this, we use FQE from d3rlpy \citep{seno2021d3rlpy}. FQE takes as input a dataset of transitions $D=\left\{s_{t}, a_{t}, s_{t+1}, r_{t}\right\}_{t=1}^{n}$ and a policy $\pi$, and, at each step $k$ of the algorithm, computes the targets $y_{t}=r_{t}+\gamma Q_{k-1}\left(s_{t+1}, \pi\left(s_{t+1}\right)\right)$ using $D$. From there, we solve $Q_k = \text{argmin}_{f \in F}\sum_{i=1}^{n}\left(f\left(s_t, a_t\right)-y_{t}\right)^{2}$ where $F$ is the function class containing all functions that can be calculated by the neural network. This outputs a neural network $Q_\pi$ which estimates the value of any state-action pair $(s,a)$ in $D$ under policy $\pi$. The performance of a policy can then be computed by taking the mean initial state value, where the initial state represents the first four hours of ventilation. Although DeepVent was trained with intermediate rewards, FQE's value estimation only depends on the dataset $\cal{D}$ and the actions chosen by the policy $\pi$ used to train FQE. Because we trained FQE using the dataset without intermediate rewards for both DeepVent- and DeepVent, the estimates are solely based on the terminal reward and can thus be used as a fair comparison. Since the physician policy effectively generates the episodes in our dataset, its discounted return for each initial state can be computed by taking the cumulative discounted reward for the episode starting at that state.

\section{Results} 

We first investigate the performance of DeepVent with FQE and compare it to physicians and BC. We then consider the safety of our choices, compared to physicians and DDQN. We further evaluate our model in OOD to show that DeepVent maintains high performance when applied to outlier patients, making it safer for real-world implementation.

\subsection{DeepVent Overall Performance}
\label{overall}

We first compare the performance of DeepVent- (CQL without intermediate reward), DeepVent (CQL with intermediate reward), the physician and behavior cloning (BC) (see Table \ref{table:FQE}) using FQE (see section \ref{OPE}), which was run for 1 million steps until convergence.

\begin{table}[h]
\fontsize{9}{10}\selectfont
\begin{center}
\begin{sc}
\resizebox{\columnwidth}{!}{
\begin{tabular}{lcccr}
Physician & BC & DeepVent- & DeepVent \\
\midrule
$\mathbf{0.502}\pm .007$ & $\mathbf{0.572 }\pm .002$ & $\mathbf{0.729}\pm .002$ & $\mathbf{0.743}\pm .005$\\
\end{tabular}
}
\end{sc}
\end{center}
\caption{Mean initial state value estimates for physician, behavior cloning (BC), DeepVent- and DeepVent, with std. errors. DeepVent- significantly outperforms both physicians and behavior cloning. Adding the Apache II intermediate reward (DeepVent) further improves the estimate.}
\label{table:FQE}
\end{table}


We observe that BC achieves a similar performance to physicians, suggesting that supervised learning can learn a relatively good policy. DeepVent- outperforms physicians by a factor of 1.45. The addition of the intermediate reward increases this factor to 1.48. Our results thus suggest that DeepVent significantly outperform both physicians and BC.

\subsection{DeepVent and Safe Recommendations}
\label{safety_section}

\begin{figure}[h!]
\centering
\centerline{
\includegraphics[width=\columnwidth]{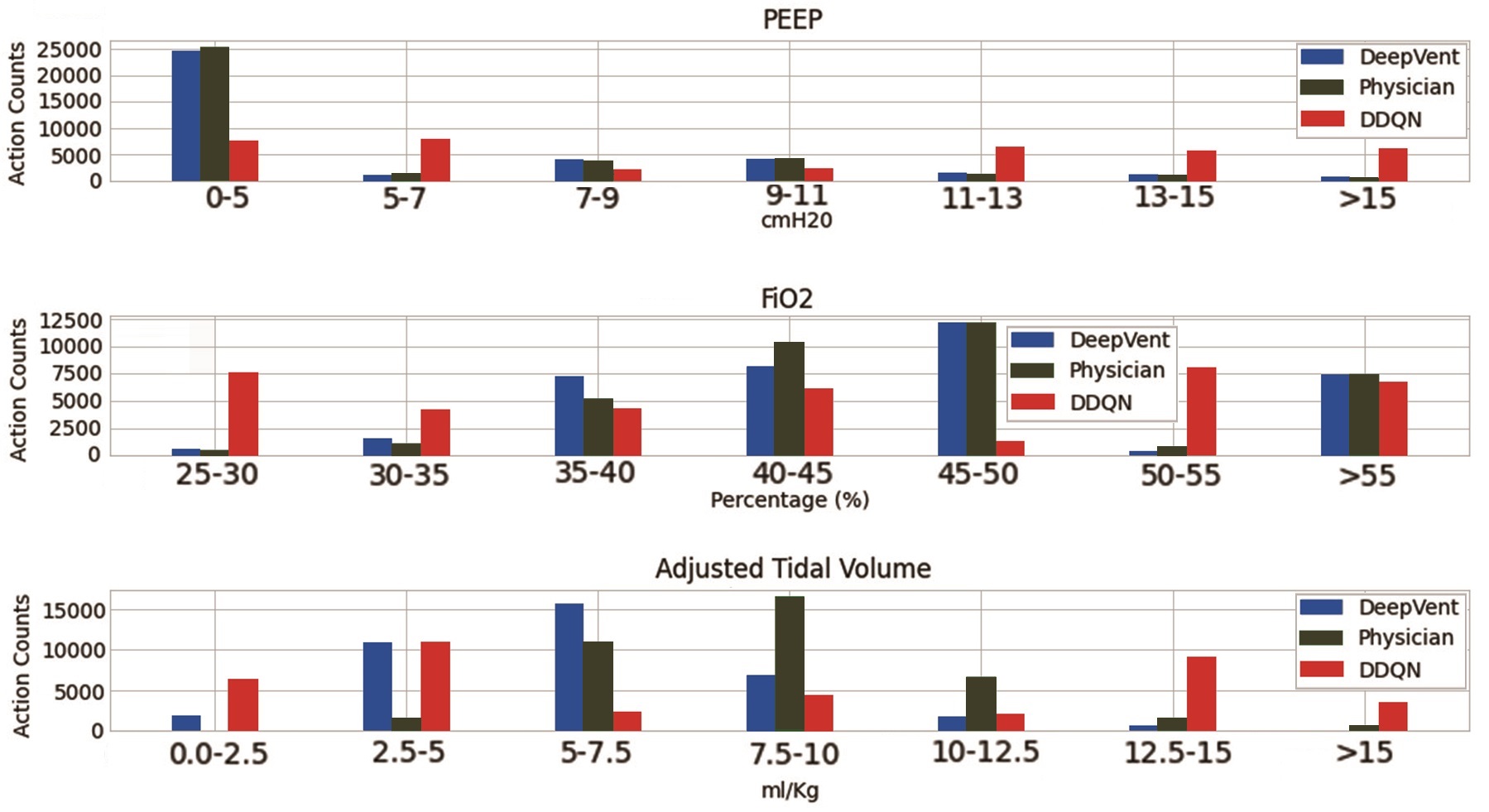}}
\caption{Distribution of actions across ventilator settings. Unlike DDQN, DeepVent makes recommendations in safe and clinically relevant ranges for each setting}
\label{fig:action_distr}
\end{figure}

We next evaluate DeepVent's action distributions (blue) compared to DDQN (red) and physicians (green) (see Figure \ref{fig:action_distr}).

The standard of care in PEEP setting is commonly initiated at 5 cmH2O \citep{nieman2017} which is supported by the high number of recommendations by physicians being in the range of 0-5 cmH2O in our dataset. DeepVent spontaneously chose to adopt this strategy by making most recommendations in the range of 0-5 cmH2O. In contrast, DDQN chose settings distributed along all the options, ranging up to 15 cmH2O, where physicians rarely went. High PEEP settings have been associated with higher incidence of complications such as  pneumothorax \citep{Zhou2021} and inflammation \citep{Guldner2016}, and should thus be avoided.

In terms of FiO2, DeepVent once again followed clinical standards of care. More specifically,  DeepVent chose actions in the same ranges as the physicians in our dataset, with many recommendations in the ranges of 35-50\% and \textgreater 55\%. In contrast, DDQN made few recommendations in these ranges, and many in ranges rarely used by physicians.

Finally, for the ideal weight adjusted tidal volume, the optimal value is usually in the 4-8 ml/kg range \citep{Luks2013,Kilickaya2013}. DeepVent made a majority of choices within 2.5-7.5 ml/kg, with most in the 5-7.5 ml/kg range. In contrast, DDQN made many recommendations in higher ranges, often above 15 ml/kg, a range rarely observed in clinical practice and associated with increased lung injury and mortality \citep{Neto2012}.

Overall, we thus observe that DeepVent, in constrast to DDQN, is able to offer safe recommendations for patients.

\subsection{DeepVent in Out-Of-Distribution (OOD)}
\begin{figure}[h!]
\begin{center}
    \includegraphics[width=0.42\columnwidth]{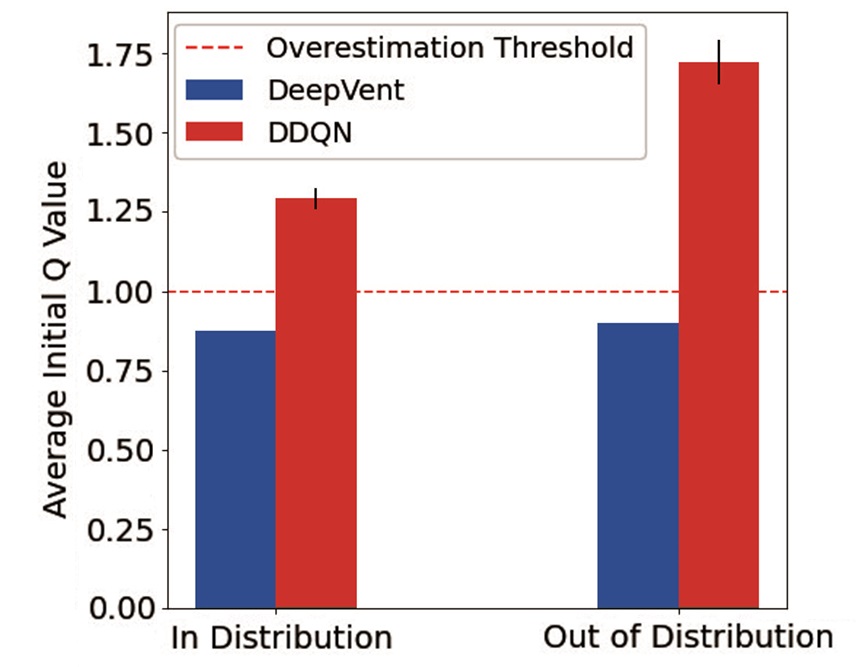}
    \caption{Mean initial Q-values for ID and OOD for DeepVent and DDQN with variances. The horizontal line is the maximum expected return. In contrast to DeepVent, DDQN clearly suffers from overestimation, aggravated when OOD}
    \label{fig:DDQN_cql_ood}
\end{center}
\end{figure}

As discussed in Sections \ref{sec:2.4}-\ref{sec:2.6}, CQL was introduced to combat the overestimation of OOD state-action pairs, a common problem in offline RL which, in the healthcare setting, can lead to dangerous recommendations.
We thus investigate whether the sub-optimal recommendations made by DDQN might be caused by overestimation of OOD states/actions. We here compute the mean initial Q values for DeepVent and DDQN estimated by FQE trained on our dataset, both in and out of distribution (see Figure \ref{fig:DDQN_cql_ood}).

\begin{figure}[h!]
\begin{center}
    \includegraphics[width=0.95\columnwidth]{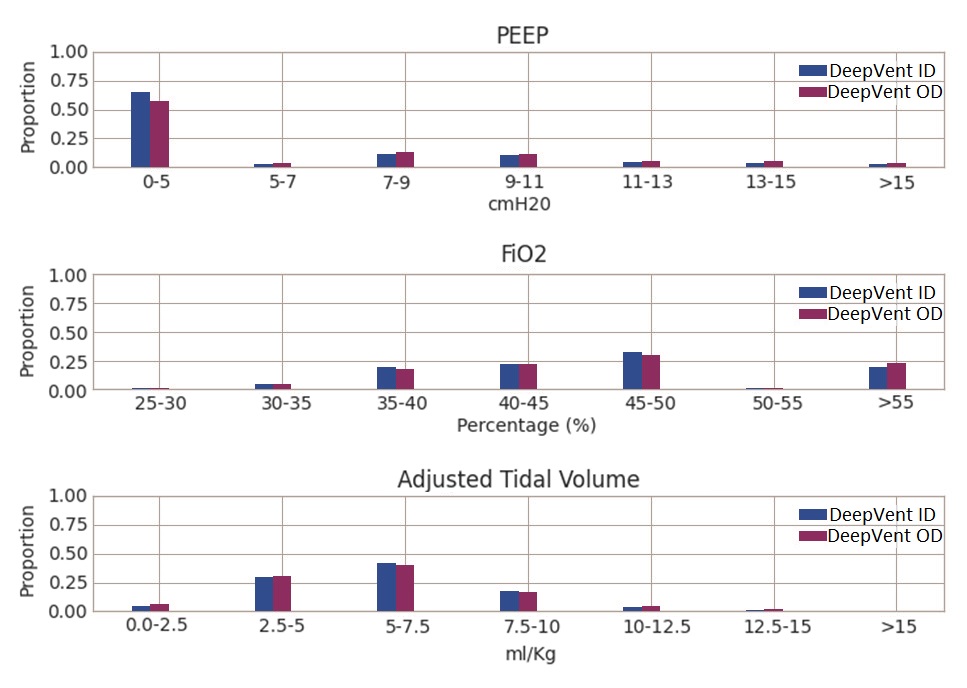}
    \caption{Distribution of actions across settings for DeepVent in distribution (ID) and out-of-distribution (OOD). DeepVent maintains settings in safe ranges in OOD data}
    \label{fig:OOD_actions}
\end{center}
\end{figure}

Since the maximal return for an episode in our data set without intermediate rewards is set at 1, and FQE was trained on this data set, values above this threshold should be considered as overestimated. We observe that DDQN overestimates values in both the ID and OOD settings. In addition, DDQN's overestimation is exacerbated in the OOD setting. This failure to accurately assess these OOD states may be the cause of the unsafe recommendations discussed above. DeepVent seems to avoid these problems, as its average initial state value estimate stays below the overestimation threshold of 1 in both settings, and barely changes in OOD, suggesting stability of the model in both settings. To strengthen this hypothesis, the action distribution of DeepVent in the OOD setting was investigated (see Figure \ref{fig:OOD_actions}).

The distribution in the OOD setting  closely resembles the one from the ID setting, with most PEEP recommendations in 0-5 cmH2O and the majority of tidal volume recommendations in the 5-7.5 ml/Kg range. Following the clinical studies outlined in Section \ref{safety_section}, DeepVent's safety in terms of recommendations extends to the OOD setting.

\section{Discussion \& Conclusion}

\paragraph{Summary}

We develop DeepVent, a decision support tool for safe mechanical ventilation treatment using offline deep reinforcement learning. We show that our use of Conservative Q-Learning leads to settings in clinically relevant and safe ranges, by addressing the overestimation of the values of out-of-distribution state-action pairs.  Furthermore, we show, using FQE, that DeepVent achieves a higher estimated performance when compared to physicians, which can be further improved by implementing our Apache II based intermediate reward. We conclude that DeepVent intuitively learns to pick actions that a physician would agree with, while using its capacity to overview vast amounts of data and understand the long-term consequences of its actions to improve outcomes for patients. Moreover, the fact that DeepVent is associated with low overestimation in out-of-distribution data makes it highly reliable, reducing the gap between research and real-world implementation.

\paragraph{Limitations}
While FQE has been shown to be a highly reliable evaluation method \cite{tang2021model}, it is important to note that our reported performance (see Section \ref{overall}) is an estimation rather than an exact value. Further works evaluating performance in a clinical setting or in simulators would permit a more reliable evaluation. Further investigation of data amputation methods could be performed to guarantee methods that mimic the protocols in ICUs. Furthermore, despite its large size and strong reputation, the MIMIC-III dataset is limited to a specific geographic location and may thus represent certain patient populations with more importance than others.

\paragraph{Future Directions}  
DeepVent is expected to significantly improve outcomes for patients under ventilation, with the potential to automatically adjust ventilator settings with high performance. At first, DeepVent could be deployed as a decision support tool, where physicians can either agree or reject its decision, permitting practically no risks for patients. This is a common process in healthcare as it gives an opportunity to learn further safety constraints before autonomous deployment. Following this clinical validation phase, DeepVent will likely naturally transition to full automation, freeing up time for physicians to focus on other components of treatment. Furthermore, our work lays a foundation not only for ventilation, but more broadly for any application of RL to healthcare. We show the potential of CQL in healthcare and introduce a broadly applicable intermediate reward based on the Apache II mortality prediction score.

\section{Ethical statement}
Implementation of DeepVent through clinical trials must, as any other technology, respect a high standard of ethical considerations. A fair subject selection must be made, by which patients enrolled in the trial represent the population DeepVent will be applied to. This includes but is not limited to accurate representation of demographics such as age, sex and ethnicity. Privacy, consent and patient confidentiality must at all times be respected. Furthermore, patient welfare must be continuously monitored to ensure optimal care.

\section{Acknowledgments}

We would like to thank Bogdan Mazoure (Mila) and Eyal de Lara (University of Toronto) for sharing constructive feedback, Adam Oberman (Mila) for their advice and Andrew Bogecho (McGill) for access to McGill RL Lab resources.

\section{Code availability}

The code for this project can be found at:\\ \url{https://github.com/FlemmingKondrup/DeepVent}

\small
\bibliography{aaai23}

\end{document}